\documentclass[letterpaper, 10 pt, conference]{ieeeconf}  % Comment this line out if you need a4paper

\usepackage{amsmath,amsfonts}
\usepackage{algorithmic}
\usepackage{array}
\usepackage[caption=false,font=normalsize,labelfont=sf,textfont=sf]{subfig}
\usepackage{textcomp}
\usepackage{stfloats}
\usepackage{pifont}
\usepackage{hyperref}
\usepackage{siunitx}

\usepackage{booktabs} 
\usepackage{tikz}
\usetikzlibrary{arrows.meta,positioning,calc,fit,decorations.pathmorphing}

\hypersetup{
	colorlinks=true, % ??????????????
	linkcolor=blue,  % ???????
	citecolor=blue, % ???????
	urlcolor=red     % URL ?????
}
\usepackage{url}
\usepackage{verbatim}
\usepackage{graphicx}
\usepackage{xcolor}
\def\BibTeX{{\rm B\kern-.05em{\sc i\kern-.025em b}\kern-.08em
		T\kern-.1667em\lower.7ex\hbox{E}\kern-.125emX}}
\usepackage{balance}

\definecolor{yhblue}{RGB}{10, 30, 150}

\IEEEoverridecommandlockouts                              % This command is only needed if 

\title{\LARGE \bf
A Unified Multi-Dynamics Framework for Perception-Oriented Modeling in Tendon-Driven Continuum Robots}

\author{Ibrahim Alsarraj$^{1}$, Yuhao Wang$^{1}$, Abdalla Swikir$^{1}$, Cesare Stefanini$^{1}$, Dezhen Song$^{1}$, Zhanchi Wang$^{2*}$, Ke Wu$^{1*}$ % <-this % stops a space
\thanks{Manuscript received November 22, 2025; Revised April 2, 2026; Accepted April 21, 2026.}%
\thanks{This paper was recommended for publication by Editor Yong-Lae Park upon evaluation of the Associate Editor and Reviewers comments.}%
\thanks{This work was supported by MBZUAI Projects \#848085.}%
\thanks{$^{1}$Robotics Department, Mohamed bin Zayed University of Artificial Intelligence (MBZUAI), Abu Dhabi, United Arab Emirates.}%
\thanks{$^{2}$Hefei National Research Center for Physical Sciences at the Microscale, University of Science and Technology of China (USTC), Hefei, China.}%
\thanks{Corresponding authors: Zhanchi Wang (zhanchi@ustc.edu.cn) and Ke Wu (ke.wu@mbzuai.ac.ae).}%
\thanks{Digital Object Identifier (DOI): see top of this page.}
}

\begin{document}

\maketitle
\thispagestyle{empty}
\pagestyle{empty}

%%%%%%%%%%%%%%%%%%%%%%%%%%%%%%%%%%%%%%%%%%%%%%%%%%%%%%%%%%%%%%%%%%%%%%%%%%%%%%%%
\begin{abstract}
%Soft continuum robots enable inherently safe and contact-rich interaction, but their perception typically depends on external sensors that increase hardware cost and limit scalability. This paper introduces a multi-dynamics modeling framework for tendon-driven continuum robots that unifies motor electrical dynamics, motor–winch dynamics, and continuum body dynamics into a closed-loop formulation. Within this framework, we design a feedforward–feedback current controller that ensures robust current regulation while maintaining a physically interpretable mapping from commanded voltage to tendon forces and robot motion. Crucially, the reconstructed motor current obtained through the coupled model serves as an intrinsic sensing channel, enabling detection of environmental contact, dynamic collisions, and even coarse object shape cues without additional tactile hardware. Experiments on a spiral-inspired tendon-driven soft arm, supported by high-fidelity simulations, validate accurate current tracking and reliable interaction inference. By bridging actuation and perception in a unified model, this work establishes motor current as a scalable pathway toward sensor-free soft robotic perception.
%s

Tendon-driven continuum robots offer intrinsically safe and contact-rich interactions owing to their kinematic redundancy and structural compliance. However, their perception often depends on external sensors, which increase hardware complexity and limit scalability. This work introduces a unified deterministic multi-dynamics modeling framework for tendon-driven continuum robotic systems, exemplified by a spiral-inspired robot named Spirob. The framework integrates motor electrical dynamics, motor–winch dynamics, and continuum robot dynamics into a coherent system model. Within this framework, motor signals such as current and angular displacement are modeled to expose the electromechanical signatures of external interactions, enabling perception grounded in intrinsic dynamics. The model captures and validates key physical behaviors of the real system, including actuation hysteresis and self-contact at motion limits. Building on this foundation, the framework is applied to environmental interaction: first for passive contact detection, verified experimentally against simulation data; then for active contact sensing, where control and perception strategies from simulation are successfully applied to the real robot; and finally for object size estimation, where a policy learned in simulation is directly deployed on hardware. The results demonstrate that the proposed framework provides a physically grounded way to interpret interaction signatures from intrinsic motor signals in tendon-driven continuum robots.
\end{abstract}


\section{INTRODUCTION}

Tendon-driven continuum robots have emerged as a promising paradigm for safe and compliant interaction in unstructured environments~\cite{webster2010design}, with applications spanning minimally invasive surgery~\cite{chikhaoui2018control}, inspection in confined spaces~\cite{hawkes2017soft}, and human–robot collaboration~\cite{muller2020nonlinear}. Their inherent compliance allows large, continuous deformations and natural distribution of contact forces, enabling interactions that rigid-link manipulators cannot achieve~\cite{marchese2014whole, jiang2021hierarchical}. However, this same compliance makes perception particularly challenging: accurate estimation of contact and interaction forces is essential for closed-loop control, yet remains a major bottleneck in tendon-driven continuum robotics~\cite{yasa2023overview}.

%Soft and continuum robots have emerged as a promising paradigm for safe, compliant interaction in unstructured environments~\cite{webster2010design, rus2015design}, with applications spanning minimally invasive surgery~\cite{burgner2015continuum}, inspection in confined spaces~\cite{hawkes2017soft}, and human–robot collaboration~\cite{muller2020nonlinear}. Their intrinsic compliance enables large deformations, distribution of contact forces, and exploitation of environmental constraints in ways that rigid-link manipulators cannot~\cite{marchese2014whole,jiang2021hierarchical}. However, this same compliance introduces a significant challenge: perception of contact and interaction forces. Accurate contact sensing is crucial for closed-loop behaviors, yet it remains a bottleneck in soft-continuum robotics~\cite{yasa2023overview}.
Most existing approaches to this sensing challenge rely on additional sensors. One strategy integrates force-sensing elements into the robot structure, such as arrays of flexible pressure sensors~\cite{homberg2019robust} or embedded strain gauges, to obtain direct contact measurements~\cite{mirzanejad2019soft}. While effective in limited settings, this approach is impractical for tendon-driven continuum robots, whose many degrees of freedom and distributed contact possibilities make full coverage costly and complex \cite{burgner2015continuum}. Moreover, the added wiring and sensing modules locally stiffen the body and restrict large deformations, undermining the robot’s inherent compliance~\cite{trivedi2008soft}. Another common strategy employs exteroceptive sensing, using cameras, structured light, or magnetic tracking to estimate shape and infer contact indirectly~\cite{kuppuswamy2020soft, sefati2018fbg}. Although effective for deformation tracking, these sensors cannot directly measure interaction forces and are highly sensitive to occlusion, lighting conditions, and magnetic disturbances \cite{yip2014model}. Their bulky setups further limit use in the confined or cluttered environments where continuum robots excel \cite{shi2016shape}.

%Most existing solutions to this sensing challenge rely on augmenting the system with additional hardware. One common strategy integrates force-sensing elements into the robot’s structure, e.g., arrays of flexible pressure sensors or embedded strain gauges, providing direct contact measurements~\cite{ homberg2019robust,teyssier2021human,mirzanejad2019soft}. While effective in some contexts, this approach is impractical for soft/continuum robots, whose high degrees of freedom and spatially distributed potential contact locations make comprehensive coverage prohibitively costly; moreover, the added wiring and modules locally stiffen the structure and impede large-strain deformations, undermining the robot’s compliant advantage~\cite{trivedi2008soft}. Another widely adopted strategy employs exteroceptive sensors, such as cameras, structured light, or magnetic tracking, to observe robot shape and infer contact indirectly~\cite{kuppuswamy2020soft,sefati2018fbg}. These afford good deformation tracking, but they struggle to measure interaction forces directly, and their performance remains vulnerable to occlusion, lighting, field disturbances or line-of-sight limitations.

As an alternative, intrinsic sensing leverages the robot’s own actuation signals for perception. For example, motor current has been exploited in rigid manipulators to realize full-body tactile awareness~\cite{iskandar2024intrinsic}. In tendon-driven continuum robots, tendons transmit both actuation and external loads, making their tension a rich source of information~\cite{oliver2019continuum}. Prior studies have employed inline load cells or distal tension sensors to infer external contacts~\cite{rubio2021new}. However, most existing methods only model continuum-robot dynamics, lacking a unified framework that connects commanded voltage, motor and winch dynamics, tendon transmission, and robot deformation leading up to interaction forces~\cite{sadati2020stiffness, renda2020geometric}.

%As an alternative, intrinsic sensing aims to leverage the robot’s own actuation channels for perception. For example, exploiting motor-current-based observers (e.g., in rigid manipulators) for full-body sense of touch in a rigid robotic arm~\cite{iskandar2024intrinsic}. Within tendon-driven soft manipulators, for instance, the tendons mediate both actuation and external load pathways, making their tension a rich information source~\cite{oliver2019continuum}. Some prior works have used inline load cells or distal tension sensors to reconstruct external contacts~\cite{zhou2024tension,rubio2021new}. However, existing approaches typically treat the electrical, mechanical and continuum‐domains separately, lacking a unified framework that links commanded voltage through motor dynamics, tendon/winch dynamics, and continuum body deformation up to interaction forces~\cite{sadati2020stiffness, renda2020geometric}.

In this paper, we address this gap by introducing a unified deterministic multi-dynamics modeling framework for tendon-driven continuum robots that integrates three coupled domains: (i) motor electrical dynamics, (ii) motor–winch dynamics, and (iii) continuum robot dynamics. To the best of our knowledge, this is the first framework to jointly model all three dynamics within a unified formulation for tendon-driven continuum robots. This end-to-end formulation establishes a clear causal link from commanded motor voltage to robot–environment interaction. 
%In this paper, we address this gap by proposing a multi-dynamics modeling framework for tendon-driven continuum robots that couples three domains: (i) motor electrical dynamics, (ii) motor–winch dynamics, and (iii) continuum robot dynamics including tendon and contact forces. This end-to-end formulation establishes a causal pathway from commanded motor voltage all the way to robot-environment interaction. Within this framework we design a feedforward–feedback current controller to robustly regulate motor current, while simultaneously reconstructing the observed motor current as an intrinsic sensing signal. Remarkably, these reconstructed currents reveal external contacts, dynamic collisions and even coarse object-shape cues, without any additional tactile hardware.
The contributions of this work are as follows:  

\begin{itemize}
    \item \textbf{Unified deterministic multi-dynamics framework:} We present a multi-dynamics modeling framework that unifies motor electrical, motor–winch, and continuum robot dynamics. It accurately reproduces electromechanical behaviors such as hysteresis, delay, and contact transients, bridging the gap left by single-dynamics models that fail to capture these temporal features.

    \item \textbf{Perception-enabled dynamic modeling:} Unlike typical robot dynamics models, the proposed framework captures the coupled electromechanical signals linking motor current, tendon force, and continuum deformation, enabling accurate reconstruction of interaction dynamics for perception tasks.
    
    \item \textbf{Sim-to-real alignment:} The framework closely matches real-system behavior and provides a reliable simulation environment for developing and validating perception or control strategies, thereby reducing experimental overhead.
\end{itemize}

\section{Problem Statement}

\begin{figure*}[t]
	\vspace*{0.2cm} 
    \centering
    % Replace with actual figure later
    \includegraphics[width=0.95\linewidth]{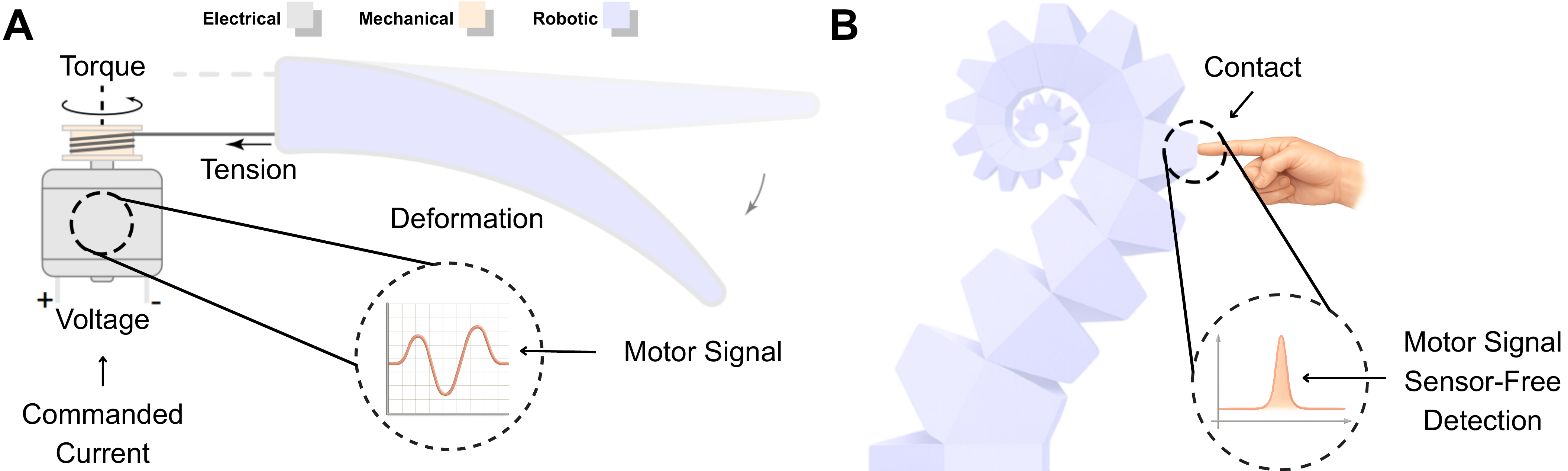}
    \caption{
    (A) Commanded motor current producing motor torque and tendon tension that drives the robot to deform. (B) Investigated Soft Robot - SpiRob.}
    \label{fig:system_overview}
\end{figure*}
\begin{figure*}[h]
	\vspace*{0.3cm} 
    \centering   \includegraphics[width=0.97\textwidth]{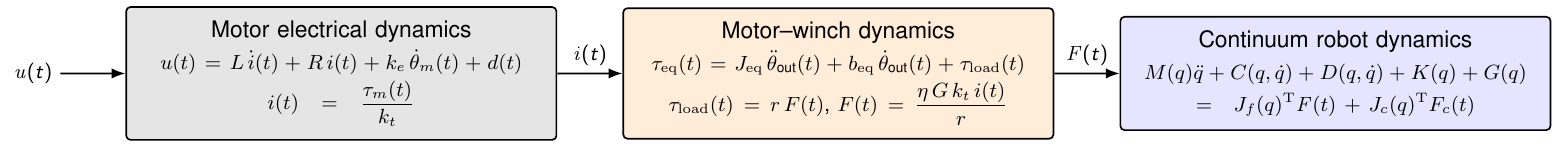}
\caption{Schematic of the multi-dynamics modeling framework.}
    \label{fig:Framework}
\end{figure*}

\subsection{The Studied Tendon-driven Continuum Robotic System}

\subsubsection{\textbf{System Architecture}}

As shown in Fig.~\ref{fig:system_overview}A, the studied tendon-driven continuum robotic system comprises three coupled subsystems: the electrical system, the motor--winch transmission, and the continuum robot.
The motor, given a commanded current and then regulated by a low-level voltage controller, drives the winch through its torque to actuate the tendons. Tendon motion deforms the continuum robot, and external contacts on the robot are transmitted through the tendons and reflected in the motor response.

\subsubsection{\textbf{The Studied Manipulator}}

%We select a highly dexterous spiral-inspired continuum robot, \textbf{SpiRob}~\cite{wang2025spirobs}, as the representative system. As shown in Fig.~\ref{fig:system_overview}, SpiRobs exhibit a natural gradient in stiffness and curvature from base to tip, enabling a wide workspace and smooth transitions between motion primitives such as reaching, wrapping, and grasping. The structure can be rapidly prototyped by 3D printing flexible filaments, while actuation is achieved with a minimal number of tendons, reducing mechanical complexity yet maintaining strong coupling between actuation and deformation, which contributes to the robot’s high dexterity.

We select a highly dexterous spiral-inspired continuum robot, \textbf{SpiRob}~\cite{wang2025spirobs}, as the representative system. 
As shown in Fig.~\ref{fig:system_overview}B, SpiRobs exhibit a natural gradient in stiffness and curvature from base to tip, enabling a wide workspace and smooth transitions among motion primitives such as reaching, wrapping, and grasping. 
The structure can be rapidly prototyped by 3D printing flexible filaments, and actuation is achieved with a minimal number of tendons, reducing mechanical complexity while preserving strong actuation–deformation coupling that underlies the robot’s high dexterity.

%This combination of high dexterity, compliance, and actuation–structure coupling provides a solid physical foundation for studying contact perception within tendon-driven continuum robots.

\subsection{Task Objective}

The objective of this work is to develop a unified deterministic multi-dynamics framework for tendon-driven continuum robots that integrates motor electrical, motor–winch, and continuum robot dynamics into a single framework. The framework reproduces key electromechanical behaviors and enables perception-oriented modeling by interpreting the signatures encoded in motor signals.

\section{Multi-dynamics Modeling Framework}
\label{sec:framework}

The studied tendon-driven continuum robotic system is modeled by combining the motor’s electrical dynamics, the motor–winch dynamics, and the continuum robot dynamics, as shown in Fig.~\ref{fig:Framework}.
We first present the dynamic models of the three subsystems and then describe how they are integrated into a unified formulation for numerical implementation (Fig. \ref{fig:closed_loop}).

\subsection{Motor Electrical Dynamics}

The electrical dynamics of a DC motor can be modeled as an equivalent $R$--$L$ circuit with a back electromotive force (EMF). The governing voltage equation is
\begin{equation}
\small
u(t) = L\,\dot{i}(t) + R\,i(t) + k_e\,\dot{\theta}_\mathrm{m}(t) + d(t),
\label{dynamics1}
\end{equation}
where $u(t)$ is the applied motor voltage, $i(t)$ is the armature current, 
$L$ and $R$ denote the motor inductance and resistance, respectively, 
$k_{\mathrm{e}}$ is the back-EMF constant, ${\theta}_\mathrm{m}(t)$ is the motor angular position and $d(t)$ is an unknown but bounded model inaccuracy term. Throughout this paper, the notation $\dot{(\cdot)}$ and $\ddot{(\cdot)}$ denote the first and second time derivatives, respectively.
Furthermore, the electromagnetic torque produced by the motor is proportional to the current and given by
\begin{equation}\label{taum}
\small
\tau_\mathrm{m}(t) = k_\mathrm{t} i(t),
\end{equation}
where $k_\mathrm{t}$ is the torque constant.

\subsection{Motor--Winch Dynamics}

The motor--winch subsystem relates the motor torque to the 
dynamics of the rotor. The kinematic relation between the motor shaft 
and the output shaft is given by
\begin{equation}\label{thetam}
\small
\theta_\mathrm{m}(t) = G \theta_{\mathrm{out}}(t), \ 
\dot{\theta}_\mathrm{m}(t) = G \dot{\theta}_{\mathrm{out}}(t), \  
\ddot{\theta}_\mathrm{m}(t) = G \ddot{\theta}_{\mathrm{out}}(t),
\end{equation}
where $G$ is the gear ratio, and 
$\theta_{\mathrm{out}}(t)$ denote the output-shaft angular position. Moreover, the dynamic equation on the winch side is
\begin{equation}
\small
\tau_{\mathrm{eq}}(t) 
= J_{\mathrm{eq}} \ddot{\theta}_{\mathrm{out}}(t) 
+ b_{\mathrm{eq}} \dot{\theta}_{\mathrm{out}}(t) 
+ \tau_{\mathrm{load}}(t),
\label{dynamics}
\end{equation}
%you need to explain what is \tau_{gearbox}; maybe also consider the switch between static friction and sliding friction
where $\tau_{\mathrm{eq}}(t)$ denotes the torque delivered through the gearbox, 
$J_{\mathrm{eq}}$ is the equivalent inertia reflected to the output side, 
$b_{\mathrm{eq}}$ is the equivalent viscous damping, 
and $\tau_{\mathrm{out}}(t)$ is the load torque exerted by the tendon. The equivalent parameters are obtained by reflecting the motor-side dynamics 
through the gear ratio $G$, such that
\begin{equation}
\small
J_{\mathrm{eq}} = G^2 J_{\mathrm{m}},\ b_{\mathrm{eq}} = G^2 b_{\mathrm{m}},
\end{equation}
where $J_{\mathrm{m}}$ and $b_{\mathrm{m}}$ represent the motor rotor inertia and viscous damping, respectively. 
The torque transmitted to the output shaft is
\begin{equation}\label{taueq}
\small
\tau_{\mathrm{eq}}(t) = \eta\,G\,\tau_{\mathrm{m}}(t),
%= \eta\,G\,k_t\,i(t)
\end{equation}
with $\eta$ denoting the transmission efficiency. Besides, the rotor rotation can be determined from the tendon displacement through the winch radius $r$ as
\begin{equation}
\small
\begin{aligned}
\theta_{\text{out}} = \frac{\Delta \ell}{r},\ 
\dot{\theta}_{\text{out}} = \frac{\Delta \dot{\ell}}{r},\ 
\ddot{\theta}_{\text{out}} = \frac{\Delta \ddot{\ell}}{r},
\end{aligned}
\label{kine}
\end{equation}
where $\Delta \ell(t)$ is the tendon displacement, which depends on the robot configuration and is formally defined in \eqref{eq:tendon_displacement}.

\subsection{Robot Dynamics}

The dynamics of Spirob is expressed in standard rigid-body form using the concept of pseudo rigid body models ~\cite{howell1996loop}, 
augmented with tendon and contact forces~\cite{wang2025spirobs}. Let $q\in\mathbb{R}^{24}$ denote 
the generalized coordinates. The equation of motion is expressed as
\begin{equation}
\small
\begin{aligned}
    M(q)\ddot q + C(q,\dot q) + &D(q,\dot q) + K(q) + G(q) \\
&= J_f(q)^{\mathrm{T}} F(t) + J_c(q)^{\mathrm{T}} F_c(t),
\end{aligned}
\label{continuumdynamics}
\end{equation}
where $M(q)$ is the generalized mass matrix, $C(q,\dot q)$ collects the 
Coriolis and centrifugal effects, $D(q,\dot q)$ is the damping matrix, $K(q)$ is the stiffness matrix, 
and $G(q)$ is the gravity vector. The vector $F \in \mathbb{R}^2$ stacks the two tendon forces, 
which act on the system through the tendon Jacobian $J_f(q)$.
The term $J_c(q)^{\mathrm{T}}F_c(t)$ accounts for external and self-contact forces. 
Finally, the tendon force $F(t)$ is related to the load torque $\tau_{\mathrm{load}}$ at the rotor by
\begin{equation}
\small
F(t) = \frac{\tau_{\mathrm{load}}(t)}{r}.
\label{tauten}
\end{equation}
Besides, each tendon’s geometric displacement is then defined as
\begin{equation}
\small
\Delta \ell_i = \ell_i(q) - \ell_i(q_0), \  i = 1,2.
\label{eq:tendon_displacement}
\end{equation}
where $\ell_i(q)$ denotes the tendon length at configuration $q$, and $\ell_i(q_0)$ its reference length at the undeformed configuration $q_0=\mathbf{0}$.
This above establishes the coupling between the motor--winch dynamics and the tendon-driven continuum robot dynamics.

\begin{figure}[t]
	\vspace*{0.1cm} 
	\centering
	\includegraphics[trim=5 0 5 0, clip, scale=0.7]{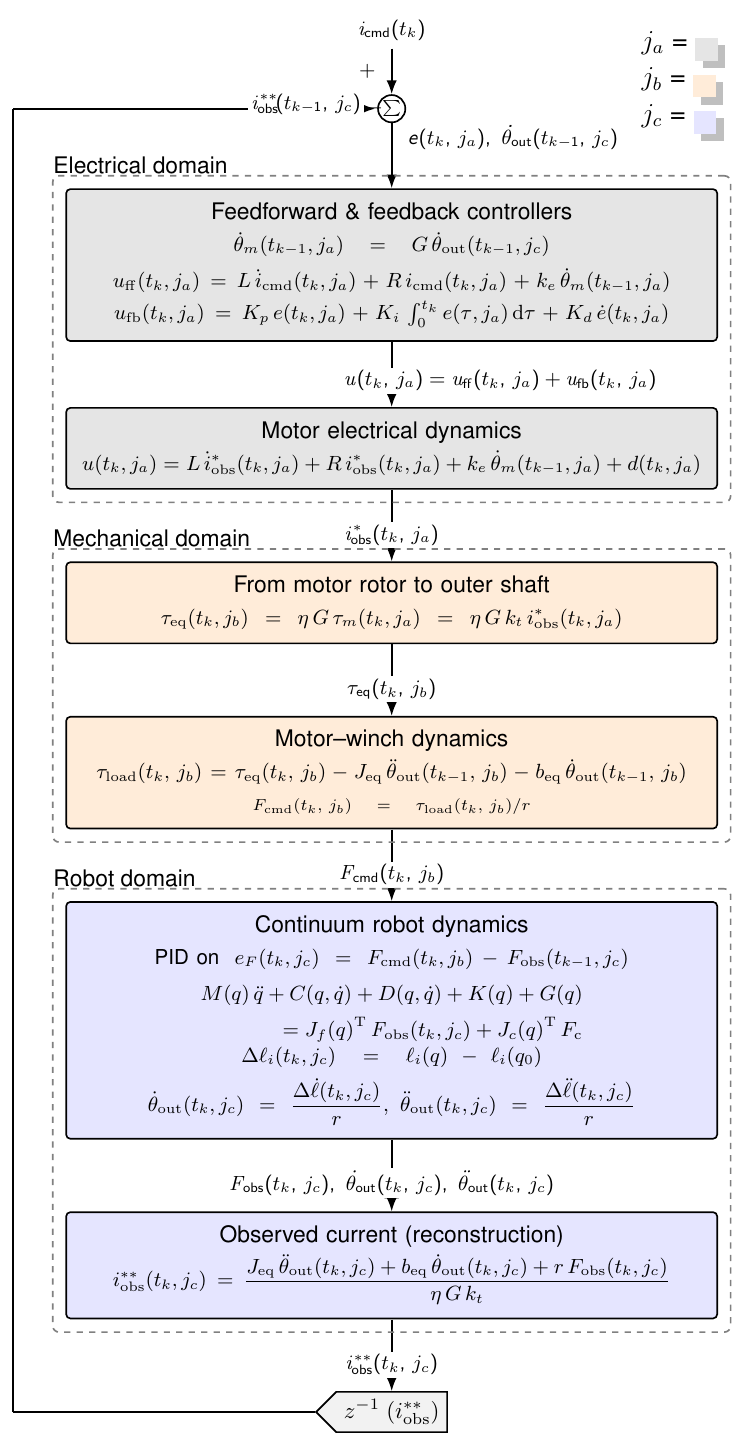}
\caption{Proposed numerical implementation of the multi-dynamics framework.}
	\label{fig:closed_loop}
%	\vspace{-3mm
\end{figure}

\subsection{System-level Coupled Flow and Implementation}
Having established the deterministic dynamic models of the three subsystems, we integrate them into a unified framework that links the commanded motor current, regulated by a low-level voltage controller, to the resulting tendon forces and robot motion.
This formulation captures the bidirectional coupling between actuation and interaction, where internal motor signals not only regulate motion through low-level feedback control but also encode information about external contacts, forming the foundation for perception-oriented modeling in tendon-driven continuum robots. As illustrated in Fig.~\ref{fig:closed_loop}, each control cycle at 
time $t_k$ proceeds in three steps:
\begin{itemize}
    \item Feedforward--feedback current control (step $j_a$)
    \item Mapping the controlled current to tendon force  (step $j_b$)
    \item Robot dynamics and current observation (step $j_c$)
\end{itemize}
We denote by $j_a$,$j_b$,$j_c$ the step counters within the $k$-th cycle. Accordingly, for identical inputs and initial conditions, the proposed first-principles framework yields the same system response.
\subsubsection{\textbf{Feedforward and feedback current control (step $j_a$)}}
At the beginning of $k$th cycle, the last available current 
$i_{\mathrm{obs}}^{\mathcal{**}}(t_{k-1},j_c)$, is known. The current error is defined as
\begin{equation}
\small
e(t_k,j_a) \;=\; i_{\mathrm{cmd}}(t_k) - i_{\mathrm{obs}}^{\mathcal{**}}(t_{k-1},j_c).
\label{eq:err_k_all}
\end{equation}
Consider the motor electrical equation with an additive disturbance as shown in \eqref{dynamics1}, we apply a composite control $u=u_{\mathrm{ff}}+u_{\mathrm{fb}}$ where the feedforward controller $u_{\mathrm{ff}}$ is derived directly from the motor electrical model \eqref{dynamics1} 
and predicts the voltage required to achieve the commanded current as
\begin{equation}
{\small
u_{\mathrm{ff}}(t_k,j_a) = L\,\dot{i}_{\mathrm{cmd}}(t_k) + R\,i_{\mathrm{cmd}}(t_k) + k_e\,\dot{\theta}_{\mathrm{m}}(t_{k-1}),
}
\label{ff}
\end{equation}
where $\dot{\theta}_m$ is obtained from~\eqref{thetam}. This anticipates the contributions of inductance, resistance, and back-EMF, thus reducing the steady-state burden on the feedback loop. The feedback controller $u_{\mathrm{fb}}$ is implemented in a PID form acting on the current error $e$ as
\begin{equation} {\small u_{\mathrm{fb}}(t_k,j_a) = K_{\mathrm{p}}\,e(t_k,j_a) + K_{\mathrm{i}}\,\underbrace{\int_{0}^{t_k} e(\tau,j_a)\,\mathrm{d}\tau}_{=:~I_e(t_k,j_a)} + K_{\mathrm{d}}\,\dot{e}(t_k,j_a) } \label{fb},\end{equation}
where $K_{\mathrm{p}}$, $K_{\mathrm{i}}$, and $K_{\mathrm{d}}$ are the proportional, integral, 
and derivative gains. This component provides robustness against unmodeled dynamics and external disturbances. Substituting $u_{\mathrm{ff}}$ \eqref{ff} and $u_{\mathrm{fb}}$ \eqref{fb} into the plant \eqref{dynamics1} and using \eqref{eq:err_k_all}
yields the closed-loop error dynamics.
Then, differentiating both sides of the resulting equation yields
\begin{equation}
\small
(L+K_{\mathrm{d}})\,\ddot e(t_k,j_a) + (R+K_{\mathrm{p}})\,\dot e(t_k,j_a) + K_{\mathrm{i}}\,e(t_k,j_a) = \dot d(t_k).
\label{eq:error-dyn}
\end{equation}
%Although the controller operates at discrete sampling instants $t_k$, 
%the sampling period $\Delta t = t_k - t_{k-1}$ is sufficiently small that the system 
%can be approximated as continuous. 
%Hence, the closed-loop error dynamics in\eqref{eq:error-dyn} 
%are analyzed in continuous time. 
As $k \to \infty$, the tracking error $e(t_k, j_a)$ converges to a small neighborhood of zero, 
whose size depends on the disturbance magnitude and the selected control gains. Then, we can update $i_{\mathrm{obs}}^*$ after the effect of the controller \eqref{ff}\eqref{fb} as follows
\begin{equation}
\small
i_{\mathrm{obs}}^{\mathcal{*}}(t_{k},j_a)=\; i_{\mathrm{cmd}}(t_k)-e(t_k,j_a),
\label{eq:err_k_all1}
\end{equation}
which drives the motor--winch subsystem.

\subsubsection{\textbf{From controlled current to commanded tendon force (step $j_b$)}}
Using the output-side dynamics \eqref{taum}\eqref{dynamics}\eqref{taueq} and the winch-tendon mapping \eqref{tauten},
the commanded tendon force at time $t_k$ is
\begin{equation}
\small
F_{\mathrm{cmd}}(t_k,j_b) =
\frac{ \tau_{\mathrm{eq}}(t_k,j_b) - J_{\mathrm{eq}}\ddot{\theta}_{\mathrm{out}}(t_{k-1},j_c) - b_{\mathrm{eq}}\dot{\theta}_{\mathrm{out}}(t_{k-1},j_c) }{r}.
\label{eq:F_cmd_tk}
\end{equation}
For implementation, $\dot{\theta}_{\mathrm{out}}$ and $\ddot{\theta}_{\mathrm{out}}$ are taken from the previous sampling instant $t_{k-1}$ and are computed using~\eqref{kine}\eqref{eq:tendon_displacement}.

\subsubsection{\textbf{Robot dynamics and current observation (step $j_c$)}}
The commanded tendon force $F_{\mathrm{cmd}}$ enters the continuum dynamics in~\eqref{continuumdynamics}. 
A feedback controller acts on the tendon-force error 
$e_F(t_k,j_c) = F_{\mathrm{cmd}}(t_k,j_b) - F_{\mathrm{obs}}(t_k,j_c)$, 
ensuring convergence as $t_k \to \infty$. 
The resulting $F_{\mathrm{obs}}(t_k,j_c)$ updates the continuum state $q(t_k)$ through~\eqref{continuumdynamics}, 
from which the rotor kinematics $\dot{\theta}_{\mathrm{out}}(t_k,j_c)$ and $\ddot{\theta}_{\mathrm{out}}(t_k,j_c)$ 
are computed using~\eqref{kine} and~\eqref{eq:tendon_displacement}. The {observed current reconstructed from the robot dynamics} is then
\begin{equation}
\small
i_{\mathrm{obs}}^{**}(t_k,j_c) \!=\!
\frac{J_{\mathrm{eq}}\,\ddot
{\theta}_{\mathrm{out}}(t_k,j_c) +b_{\mathrm{eq}}\,\dot{\theta}_{\mathrm{out}}(t_k,j_c) + r\,F_{\mathrm{obs}}(t_k,j_c)}
{\eta\,G\,k_{\mathrm{t}}}.
\label{eq:i_obs_tk}
\end{equation}
The quantity $i_{\mathrm{obs}}^{**}(t_k,j_c)$ is passed to the next cycle $k+1$ and used directly in \eqref{eq:err_k_all} at $t_{k+1}$ to close the loop (Fig. \ref{fig:closed_loop}). %This ensures consistency between $i_{\mathrm{obs}}^{**}(t)$ and $i_{\mathrm{cmd}}(t)$ despite model uncertainties.
%\Ke{cite \eqref{eq:tendon_displacement}}

\subsubsection{\textbf{Summary of one control cycle}}
As shown in Fig. \ref{fig:closed_loop}, at each sampling instant $t_k$, the closed loop proceeds in three causal steps:
\begin{itemize}
  \item Step $j_a$: Compute the current error and apply the feedforward--feedback law to obtain 
  $i_{\mathrm{obs}}^{\mathcal{*}}(t_k,j_a)$.
  \item Step $j_b$: Map $i_{\mathrm{obs}}^{\mathcal{*}}(t_k,j_a)$ to the commanded tendon force 
  $F_{\mathrm{cmd}}(t_k,j_b)$ using 
  $\dot{\theta}_{\mathrm{out}}(t_{k-1},j_c), \ddot{\theta}_{\mathrm{out}}(t_{k-1},j_c)$ from the $j-1$th cycle.
  \item Step $j_c$: Propagate $F_{\mathrm{cmd}}(t_k, j_b)$ through the robot dynamics to update
${\dot{\theta}_{\mathrm{out}}(t_k, j_c), \ddot{\theta}_{\mathrm{out}}(t_k, j_c), F_{\mathrm{obs}}(t_k, j_c)}$
and reconstruct $i_{\mathrm{obs}}^{**}(t_k, j_c)$.
The reconstructed current is then fed back to step $j_a$ at time $t_{k+1}$,
where it is compared with the new command $i_{\mathrm{cmd}}(t_{k+1})$ to close the loop.
\end{itemize}
Note that two observed currents are distinguished:
$i_{\mathrm{obs}}^{\mathcal{*}}$ refers to the current after the feedforward–feedback controller within the current loop,
whereas $i_{\mathrm{obs}}^{\mathcal{**}}$ represents the reconstructed current after propagation through the full motor–winch–robot loop.

\subsubsection{\textbf{Alternative feedback points}}
While the presented architecture regulates motor current, 
alternative feedback points are possible depending on control objectives 
and sensor availability. Since the tendon kinematics satisfy \eqref{kine} and the motor--winch--robot 
dynamics provide observable tendon force $F_{\mathrm{obs}}$, 
error functions can be defined directly as
\begin{equation}\label{fbcmd}
\small
\begin{aligned}
e_{\Delta\ell}(t_k) &=\Delta \ell_{\mathrm{cmd}}(t_k) - \Delta \ell_{\mathrm{obs}}(t_k),\\
e_{\Delta\dot\ell}(t_k) &= \Delta \dot{\ell}_{\mathrm{cmd}}(t_k) - \Delta \dot{\ell}_{\mathrm{obs}}(t_k),\\
e_F(t_k) &= F_{\mathrm{cmd}}(t_k) - F_{\mathrm{obs}}(t_k).
\end{aligned}
\end{equation}
Accordingly, the commanded current becomes a function of these errors as
\begin{equation}\label{fb}
\small
i_{\mathrm{cmd}}(t_k) = \Phi\!\left(e_\ell(t_k), e_{\dot\ell}(t_k), e_F(t_k)\right),
\end{equation}
where $\Phi(\cdot)$ denotes the feedback law mapping mechanical-domain errors 
into the current domain. The proposed framework accommodates different feedback levels, including current, displacement/velocity, or tendon force that are all consistently mapped to $i_{\mathrm{cmd}}$ through the motor dynamics. %\Ke{Did we use this the rest of the paper?}

\section{Multi-Dynamics Simulation and Validation}

To validate the proposed deterministic multi-dynamics modeling framework, we conducted physical experiments to assess accuracy across the coupled electrical, mechanical, and robot domains. The physical robot experiments provide ground-truth electromechanical responses (Fig.~\ref{fig:hardware}A), while the simulation implemented directly from the unified framework produces corresponding model predictions for one-to-one comparison (Fig.~\ref{fig:hardware}B). The following subsections detail the experimental setup, the sim-to-real parameter identification procedure, and the resulting validation of model fidelity.

\subsection{Experimental Setup}

\begin{figure}[t]
	\vspace*{0.3cm} 
    \centering\includegraphics[width=0.48\textwidth]{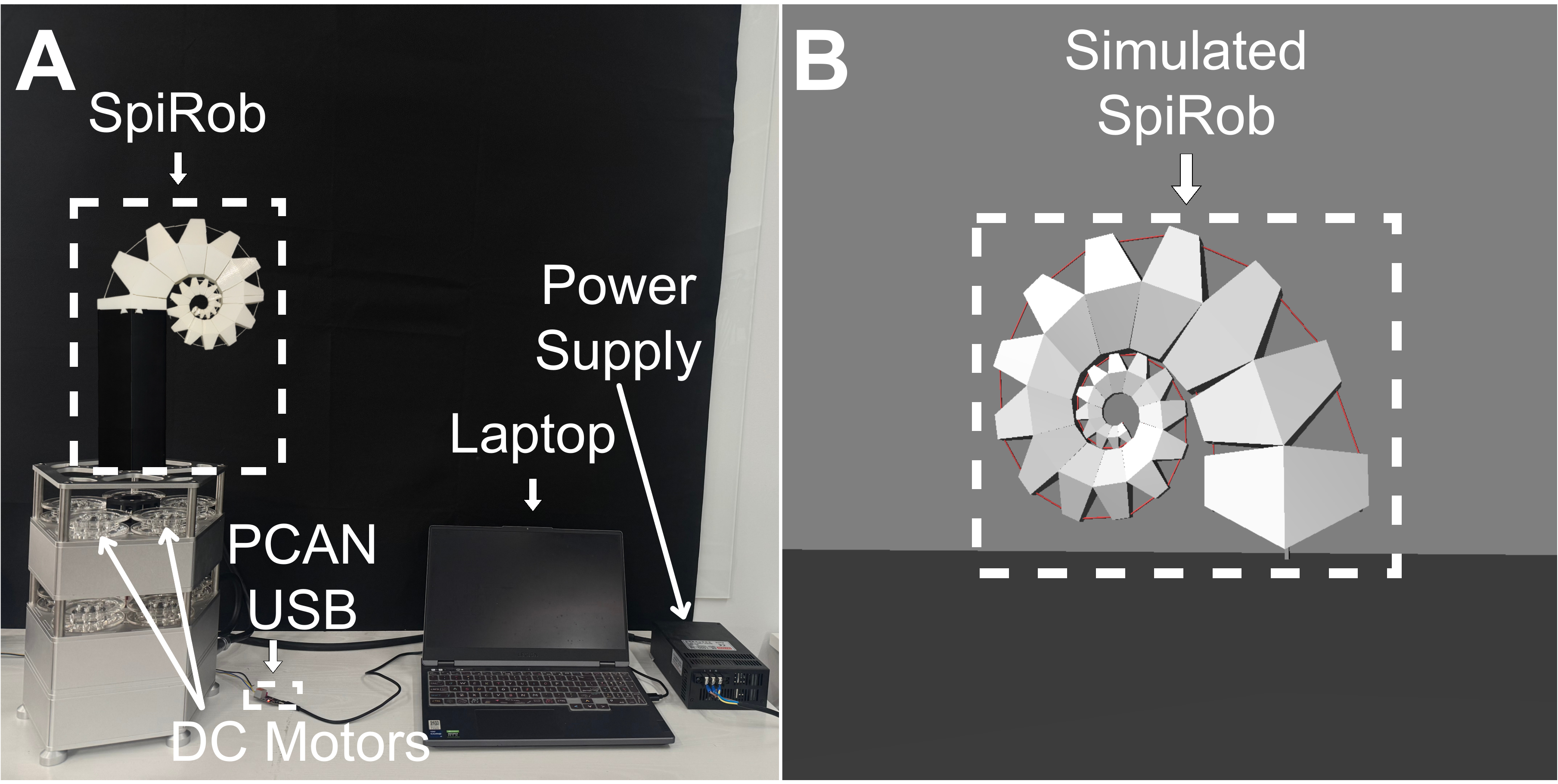}
\caption{Platform setup: (A) Experimental continuum robotic system; (B) Simulated counterpart via multi-dynamics modeling framework.}
    \label{fig:hardware}
\end{figure}

The experimental platform (Fig.~\ref{fig:hardware}A) consists of a SpiRob actuated by two DC motors through symmetrically routed tendons wound on winches.
Each motor is powered by a regulated DC source and interfaced with the laptop via a PCAN-USB adapter, which provides a CAN bus communication link for sending motor commands and receiving feedback under a \SI{1}{kHz} real-time loop.
Motor currents were recorded at full sampling rate and smoothed using a simple filter (100-sample moving average) to reduce electrical noise.

\subsection{Sim-to-Real Identification}
Key parameters in the motor-winch subsystem of the multi-dynamics simulation have been identified to align physical behaviors of the studied continuum robotic system (Fig.~\ref{fig:hardware}A). We identify these parameters by minimizing the mismatch between the modeled
and physically measured motor currents over samples $\{t_k\}_{k=1}^N$.
Let $\mathbf{p}=\big[\eta,\;b_m,\;J_m\big]^{\mathbf{T}}$
\begin{equation}
\label{eq:ident}
\begin{aligned}
    \small
\mathbf{p}^\star = &\arg\min_{\mathbf{p}\in\mathcal{P}}
\sum_{k=1}^{N}\rho\!\left(i_{\mathrm{obs}}^{**}(t_k;\mathbf{p})-i_{\mathrm{real}}(t_k)\right)^2,\\
&\textbf{s.t.}\ \eqref{dynamics1}\eqref{dynamics}
\eqref{continuumdynamics}...
\end{aligned}
\end{equation}
where
\begin{equation}
\small
\begin{split}
\mathcal{P} := \bigl\{\,\mathbf{p}:\;
& 0.6 < \eta \le 1.0,\;
5\times10^{-6} < J_m < 5\times10^{-4},\\
& 10^{-5} < b_m < 10^{-3}\bigr\}.
\end{split}
\end{equation}
\noindent
Here $i_{\mathrm{real}}(t_k)$ is the measured current from experiments sampled at time instants.
The vector $\mathbf{p}^\star$ denotes the optimal parameter set within the feasible
domain $\mathcal{P}$. Moreover, a geared transmission ($G \approx 19$) is used by default to capture reflected inertia and damping in~\eqref{dynamics}–\eqref{tauten}.
%The parameters $\kappa$ and $\mu$ represent vectors of effective stiffness and damping 
\subsection{Experimental Validation}
The experimental validation verifies that the proposed multi-dynamics framework accurately reproduces the coupled electromechanical behavior of the tendon-driven continuum robotic system and captures key nonlinearities such as hysteresis, delay, and limit-induced effects.
\begin{figure}[t]
	\vspace*{0.3cm} 
\centering
\includegraphics[width=0.48\textwidth]{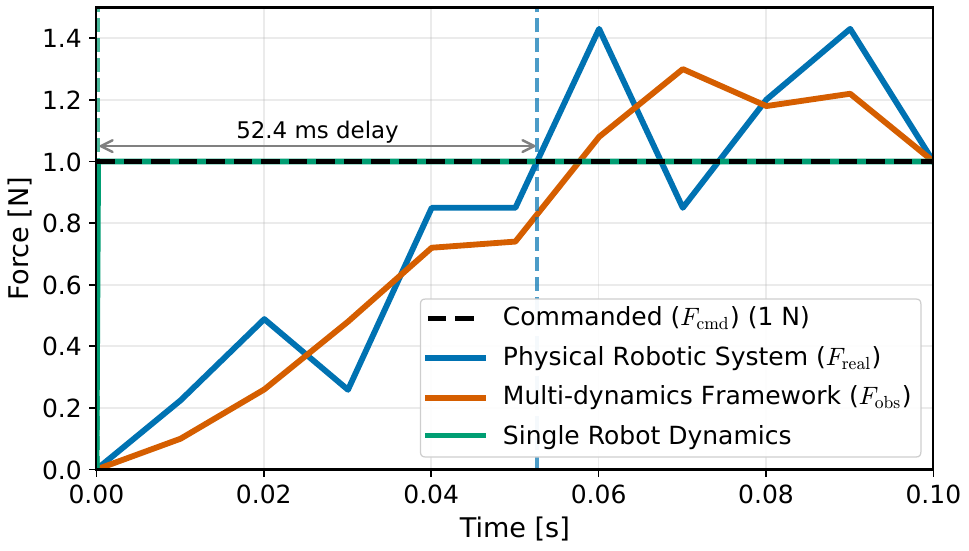}
\caption{Hysteresis and delay validation.}
\label{fig:force_regulation}
\end{figure}
\begin{figure}[t]
    \centering
    \includegraphics[width=0.43\textwidth]{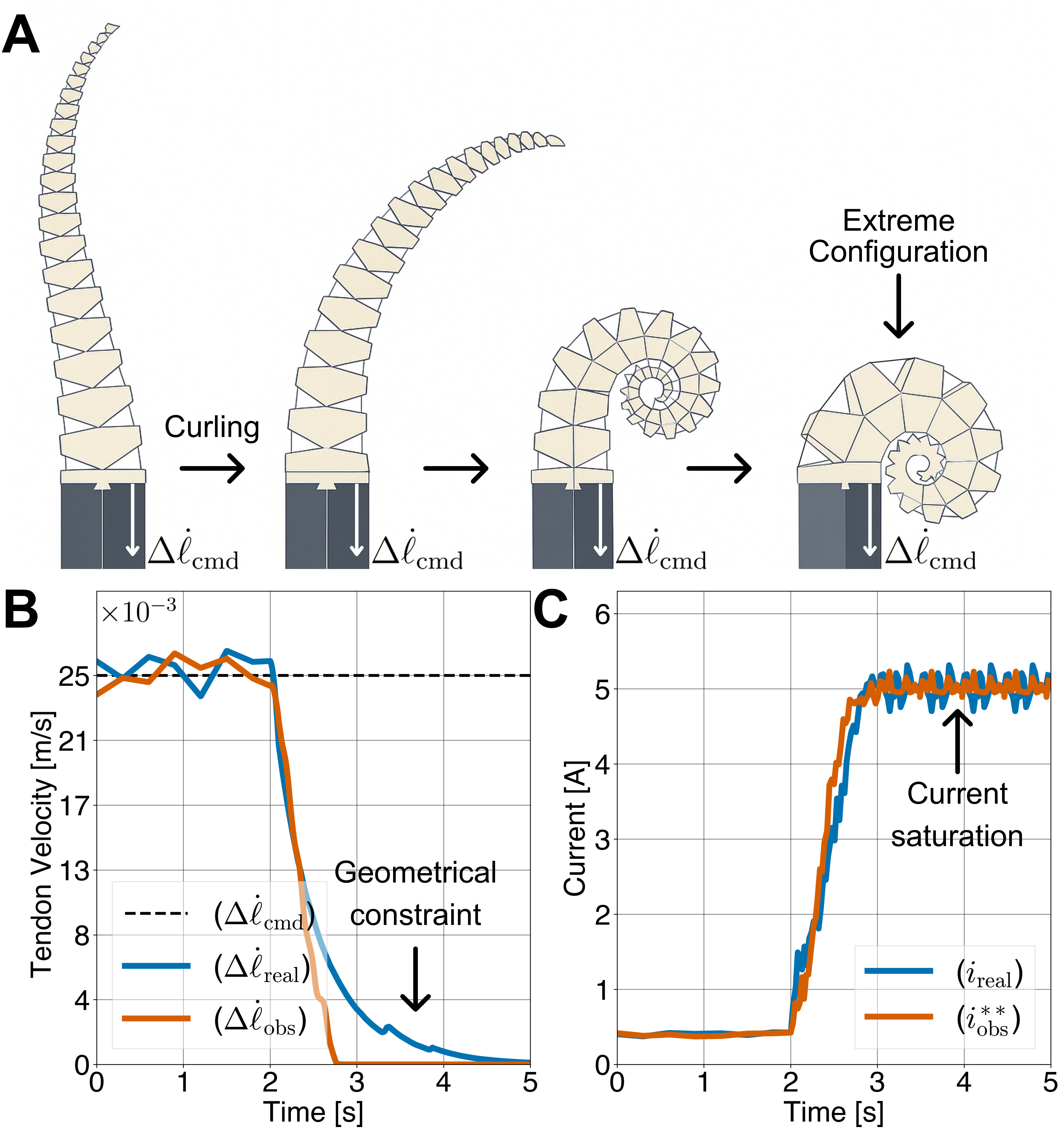}
    \caption{{Comparison of experimental and simulated motor responses during extreme curling.}}
    \label{fig:Extreme}
\end{figure}
\subsubsection{\textbf{Hysteresis and Delay}}
%This experiment assesses whether the proposed multi-dynamics framework reproduces the 
This experiment evaluates whether the proposed multi-dynamics framework can reproduce the hysteresis and delay observed in tendon force regulation of the complete system. 
The hysteresis arises from the physical processes represented by the three coupled dynamics: motor electrical, motor-winch transmission, and continuum body dynamics. 
A constant tendon force of \(1~\mathrm{N}\) was applied to the physical robot, the proposed model via \eqref{fb}, and a baseline single robot-dynamics model. As shown in Fig.~\ref{fig:force_regulation}, the physical robot (blue) exhibits rise-time lag and a steady-state delay of \(52.4~\mathrm{ms}\), both captured by the proposed framework (orange), while the single robot-dynamics model (green) responds instantaneously. This agreement is maintained under different commanded forces, confirming the framework’s accuracy across operating conditions.

%The slight delay in simulation results from the causal implementation of step~$j_b$ in~\eqref{eq:F_cmd_tk}, which uses $\dot{\theta}_{out}$ and $\ddot{\theta}_{out}$ from the previous cycle.

\subsubsection{\textbf{Extreme Configuration Detection}}

\label{sec:extreme-detection}
%To further assess model fidelity, we evaluated whether the framework reproduces the 

To further assess model fidelity, we evaluated whether the framework can reproduce the electromechanical signatures associated with the robot reaching its {extreme curled configuration}, where self-contact occurs. 
One tendon was driven at a constant commanded tendon velocity $\Delta \dot{\ell}_{\mathrm{cmd}}$ via a feedback controller~\eqref{fbcmd}, causing the body to curl until motion ceased (Fig.~\ref{fig:Extreme}A). 
Motor current reconstructed from~\eqref{eq:i_obs_tk} and tendon velocity were recorded.
During free curling, both signals remained steady; near the motion limit, tendon tension produced a sharp rise in motor current (Fig.~\ref{fig:Extreme}C) accompanied by a corresponding drop in tendon velocity (Fig.~\ref{fig:Extreme}B). 
The close agreement between experimental and simulation confirms that the proposed multi-dynamics framework accurately captures limit-induced interaction behavior, including the characteristic rise to the motor’s current saturation level approximately
\(|i_{\mathrm{obs}}^{**}|,\; |i_{\mathrm{real}}| \le 5~\mathrm{A}\)  (Fig.~\ref{fig:Extreme}C).

\section{Perception-Oriented Modeling Applications}

%%%%%%%%%%%%%%%%%%%%%%%%%%%%%%%%%%%%%%%%%%%%%%%%%%%%%%%%%%%%%%%%%%%%%%%%%%%%%%%%

\begin{figure*}[t]
	\vspace*{0.3cm} 
    \centering  \includegraphics[width=0.93\textwidth]{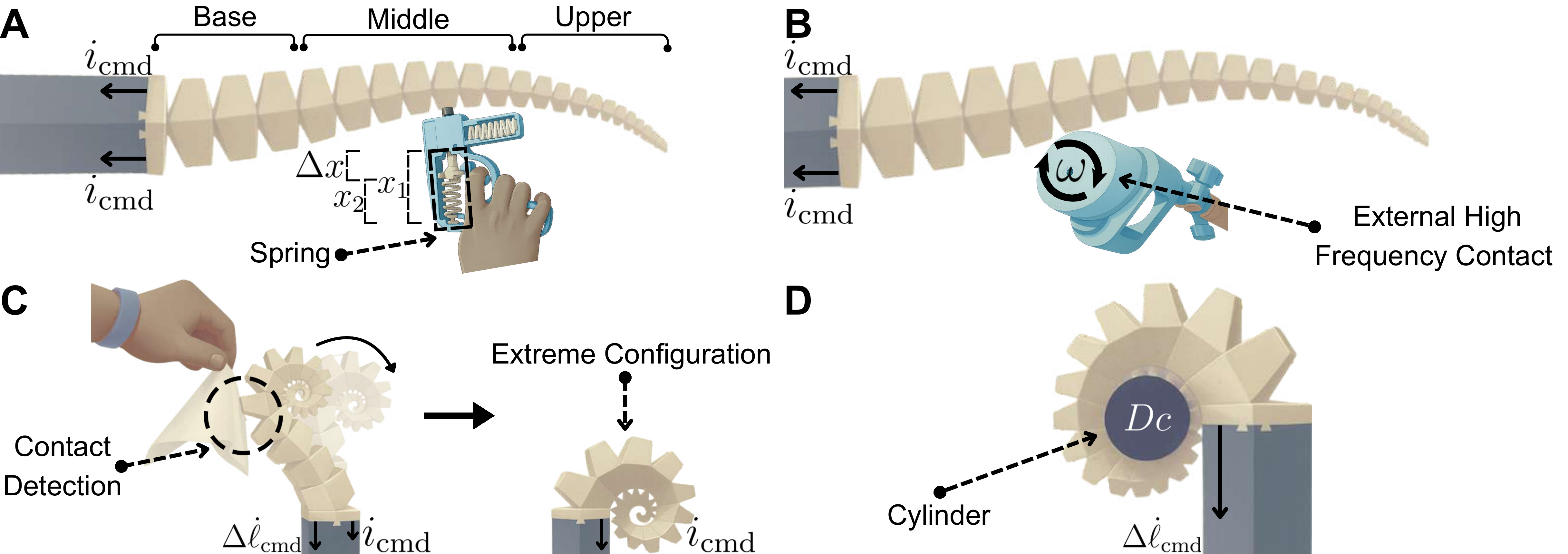}
    \caption{{Evaluation scenarios: (A) Passive perception under constant current; (B) Passive perception with high-frequency contacts; (C) Active perception for contact detection during uncurling; (D) Object-size estimation through wrapping around targets of varying diameters.}}
    \label{fig:setup}
\end{figure*}

\begin{figure*}[t]
    \centering
    \includegraphics[width=0.97\textwidth]{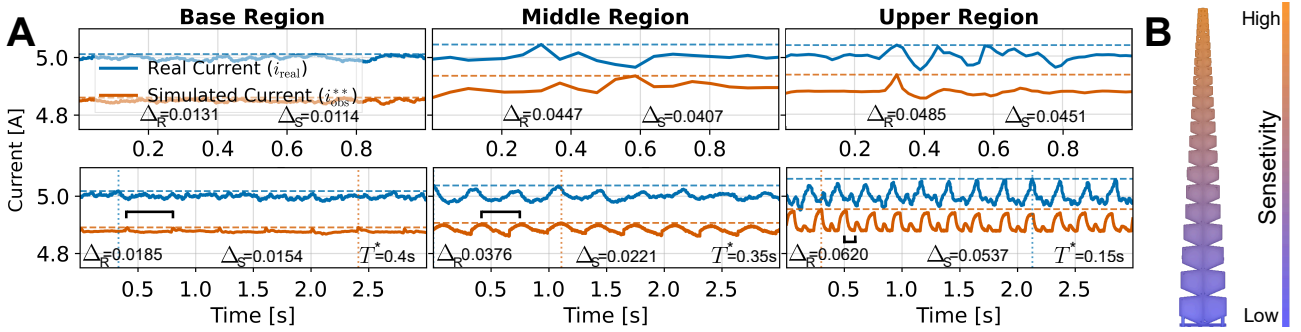}
\caption{{(A) Real and simulated motor-current responses under single (top) and periodic (bottom) contacts at different body regions. (B) Sensitivity distribution along the robot.}}
    \label{fig:passive}
\end{figure*}

To demonstrate the broader capability of the framework, we conducted experiments ranging from passive to active perception and object-size estimation, verifying passive contact sensing, validating model-based active perception and control strategies on the real robot, and demonstrating deployment of a simulation-trained policy on hardware.

%\begingroup
%\color{yhblue}

\subsection{Passive Perception}

We evaluated the framework’s ability to infer external interactions from intrinsic motor signal responses.
The robot was driven with a constant commanded current $i_\mathrm{cmd}$ applied to two tendons under two different external contact conditions (Fig.~\ref{fig:setup}A–B).
Both single-contact and periodic-contact tests were performed in the direct-drive configuration without the gear box ($G = 1$).

\subsubsection{\textbf{Single Contact Perception}}\label{scp}

A compliant spring gun was used to apply consistent contact disturbances at different locations along the robot body (Fig.~\ref{fig:setup}A). 
When released, the spring’s elastic potential energy $(\tfrac{1}{2}k_s\Delta x^2)$ was converted into the kinetic energy of a small spherical projectile $(\tfrac{1}{2}m_{\mathrm{eff}}v^2)$, 
where $k_s$ is the spring stiffness, $m_{\mathrm{eff}}$ is the projectile mass, and $\Delta x$ is the compression from equilibrium. 
To ensure identical impact conditions, the projectile velocity was determined from the energy balance as $v = \sqrt{\frac{k_s}{m_{\mathrm{eff}}}}\,\Delta x$ so that each collision occurred at the same impact speed.

According to~\eqref{continuumdynamics}, the external contact term $J_c(q)^{\mathrm{T}}F_c$ perturbs the configuration $q$ through the configuration-dependent stiffness $K(q)$, damping $D(q,\dot q)$, and mass $M(q)$.
These effects, in turn, influence the tendon forces $F_{\mathrm{obs}}$ and the tendon displacements $\Delta \ell_i(q)$ ($i=1,2$), which generate the corresponding motor rotations $\theta_{\mathrm{out}}$ via~\eqref{kine} and ~\eqref{eq:tendon_displacement}.
The observed motor current reconstructed by~\eqref{eq:i_obs_tk} thus reflects these local mechanical variations.
As shown in Fig.~\ref{fig:passive}A (top), the steady current shift increases with contact distance from the base, being negligible proximally and maximal distally.
This spatial trend aligns with the decreasing stiffness and mass distributions in $K(q)$ and $M(q)$.
Both the reconstructed current $i_{\mathrm{obs}}^{\mathcal{**}}$ and the measured current $i_{\mathrm{real}}$ exhibit smaller amplitudes and slower variations near the base, confirming that current responses encode the robot’s passive compliance profile.

\subsubsection{\textbf{Periodic Contact Perception}}

A rotating object imposed high-frequency periodic contact forces in~\eqref{continuumdynamics} at a constant angular speed~$\omega$ (Fig.~\ref{fig:setup}B), yielding a contact period $T=2\pi/\omega=0.1\,\mathrm{s}$. As shown in Fig.~\ref{fig:passive}A (bottom), the reconstructed currents $i_{\mathrm{obs}}^{\mathcal{**}}$ exhibit apparent (measured) contact periods of $T^*\approx0.4$, $0.35$, and $0.15~\mathrm{s}$ from base to tip, indicating that distal regions more closely follow the true excitation. Similar to the single-contact case in~\ref{scp}, both amplitude and frequency fidelity improve along the arm due to the decreasing stiffness and inertia in~\eqref{continuumdynamics}, which strengthen the dynamic coupling between the contact force $F_c$ and the observed current~\eqref{eq:i_obs_tk}. Consequently, the distal section exhibits higher sensitivity under high-frequency excitation (Fig.~\ref{fig:passive}B).
%A rotating object imposed periodic contact forces $F_c$ in~\eqref{continuumdynamics} at 
\begin{figure}[t]
	\vspace*{0.3cm} 
    \centering   \includegraphics[width=0.45\textwidth]{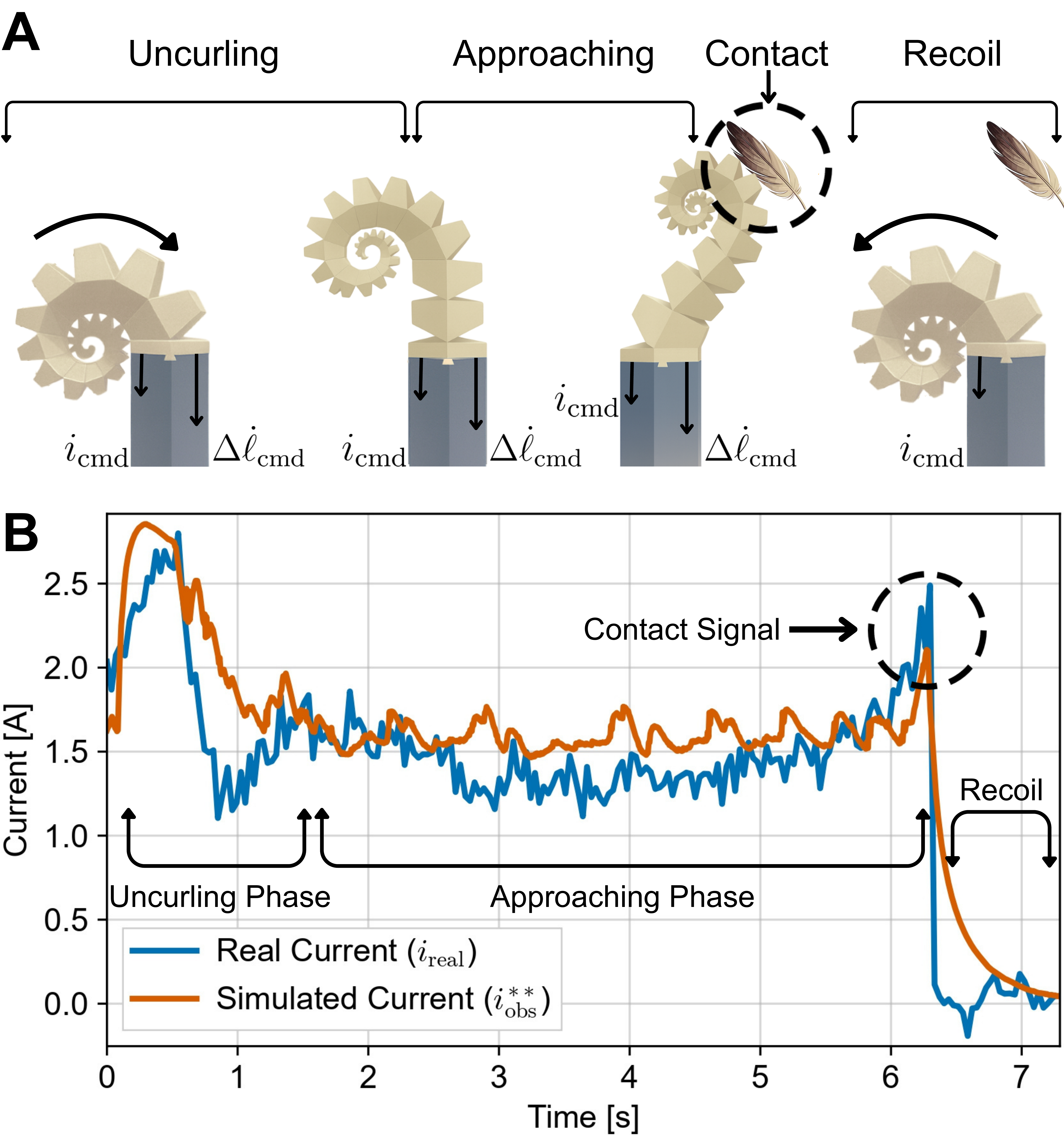}
    \caption{{Comparison of real and simulated motor currents during active perception.}}
    \label{fig:active_contact}
\end{figure}
\subsection{Active Perception}
We next evaluated the capability of the proposed multi-dynamics framework for active contact detection (Fig.~\ref{fig:setup}C and Fig.~\ref{fig:active_contact}A). The right tendon was driven at a constant tendon velocity via~\eqref{fb}, while the left tendon maintained a small constant current for baseline tension.
Starting from an extremely curled configuration, the robot uncurled until contacting an object, identified by a sharp rise in the measured motor current, before recoiling to its initial shape (Fig.~\ref{fig:active_contact}A).
Contact detection was implemented using the rate-based current observer~\eqref{eq:i_obs_tk} with a rolling-window derivative, and a contact is declared when any of the following conditions is met: an absolute rise $> 0.8~\mathrm{A}$, a relative increase $> 50\%$ of baseline, or a slope $> 6~\mathrm{A/s}$.
This adaptive rule provided robust detection across varying baseline loads and contact locations without requiring fixed thresholds \cite{wang2025spirobs}. As shown in Fig.~\ref{fig:active_contact}B, the experimental results under identical conditions closely match the simulation with an RMSE of $0.307\,\mathrm{A}$, both exhibiting gradual current evolution during free motion followed by a sharp transient upon contact.

\subsection{Object Size Estimation}
\begin{figure}[t]
	\vspace*{0.3cm} 
    \centering   \includegraphics[width=0.48\textwidth]{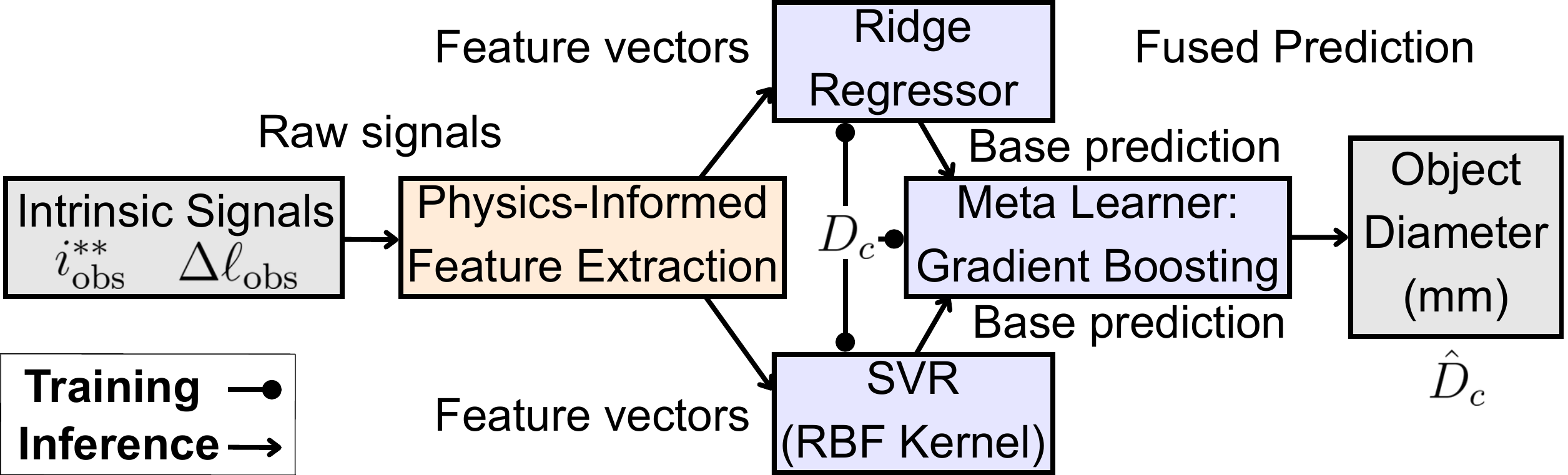}
    \caption{Ensemble learning pipeline.}
    \label{fig:Ensemble}
\end{figure}

\begin{figure}[t]
    \centering   \includegraphics[width=0.48\textwidth]{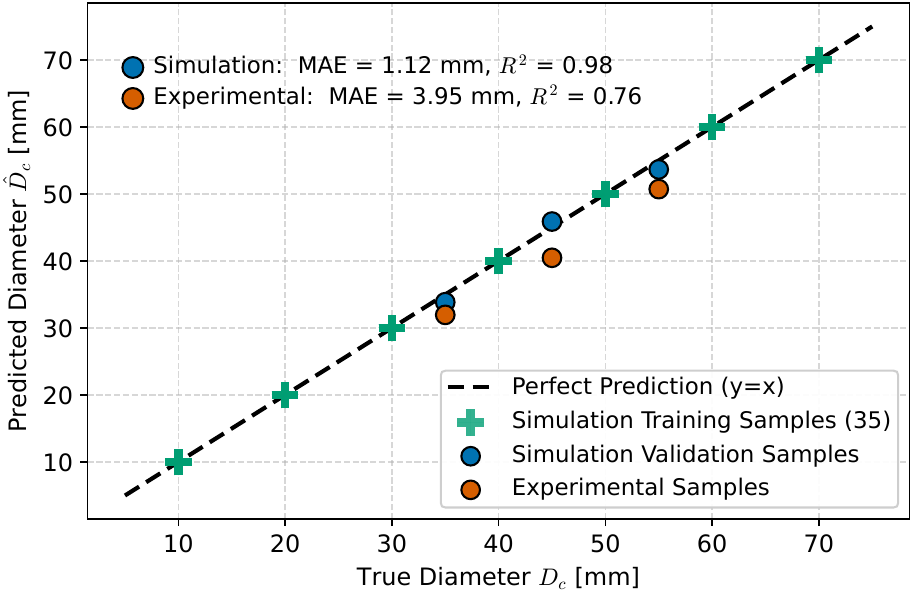}
    \caption{Comparison of predicted and true object diameters for simulation and experimental samples.}
    \label{fig:Shape}
\end{figure}

We further evaluated the framework’s ability to perform high-level, model-driven inference. The robot actively wrapped around cylinders of different diameters. The tendon was driven at a constant commanded tendon velocity $\Delta \dot{\ell}_{\mathrm{cmd}}$
 and the other left unactuated as shown in Fig.~\ref{fig:setup}D.

While deep neural networks could be employed for this task, their performance typically depends on large datasets. Instead, we employ a lightweight, interpretable ensemble learning framework that couples physics-informed feature extraction with classical regressors (Fig.~\ref{fig:Ensemble}). During training, intrinsic signals ${i_{\mathrm{obs}}^{\mathcal{**}}},\;{\Delta\ell_{\mathrm{obs}}}$ are transformed into feature vectors capturing temporal, spectral, and electromechanical characteristics. Here, the reconstructed current $i_{\mathrm{obs}}^{\mathcal{**}}$ encodes the electromechanical signatures of contact and object interaction (Fig. \ref{fig:active_contact}), while the tendon displacement $\Delta\ell_{\mathrm{obs}}$ captures the geometric state of the continuum robot during wrapping (Fig. \ref{fig:Extreme}B).
These are used to train a Ridge regressor and a support vector regressor (SVR) with a radial basis function (RBF) kernel as the base models, whose out-of-fold predictions form the input to a Gradient Boosting meta-learner. During inference, new intrinsic signals pass through the same feature extraction pipeline, and the meta-learner fuses the base predictions to produce the final object-diameter estimate $\hat{D}_c$.
A dataset of 35 simulation samples with cylinder diameters $D_c$ ranging from 10~mm to 70~mm in 10~mm increments was used to train the ensemble regression model. The model performance was evaluated using the mean absolute error (MAE), which quantifies the average absolute prediction error in millimeters, and the coefficient of determination ($R^2$), which indicates the proportion of variance in the true diameters explained by the regression model, defined as
\begin{equation}
\small
\mathrm{MAE} = \tfrac{1}{N}\sum_{i=1}^{N}\!\left|D_{c,i} - \hat{D}_{c,i}\right|, \ 
R^2 = 1 - \tfrac{\sum_i (D_{c,i} - \hat{D}_{c,i})^2}{\sum_i (D_{c,i} - \bar{D}_c)^2}
\label{eq:mae_r2},
\end{equation}
where $D_{c,i}$ and $\hat{D}_{c,i}$ denote the true and predicted cylinder diameters for the $i$-th sample, and $\bar{D}_c$ is their mean. As shown in Fig.~\ref{fig:Shape}, the model achieved high accuracy in simulation for new unseen validation samples of varying sizes. ($\mathrm{MAE}=1.12$~mm, $R^2=0.98$) and maintained strong generalization to experimental samples without retraining ($\mathrm{MAE}=3.95$~mm, $R^2=0.76$).

\section{CONCLUSIONS}
This letter presents a unified deterministic multi-dynamics framework for tendon-driven continuum robots that couples motor electrical, motor–winch, and continuum robot dynamics, establishing motor signals as a intrinsic sensing channels. Experiments on a spiral-inspired robot validated the model’s fidelity and strong sim-to-real alignment, capturing hysteresis, delay, and contact-induced current responses, and demonstrated perception-oriented modeling across passive, active, and object-size estimation tasks. The results confirm that the proposed framework enables scalable, physically grounded integration of actuation and perception for soft robots, reducing reliance on external tactile hardware. Future work will extend the framework to full state estimation, enhanced perception, and probabilistic modeling for uncertainty in dynamic and varying environments.
\addtolength{\textheight}{-12cm}   % This command serves to balance the column lengths

% %%%%%%%%%%%%%%%%%%%%%%%%%%%%%%%%%%%%%%%%%%%%%%%%%%%%%%%%%%%%%%%%%%%%%%%%%%%%%%%%

\bibliographystyle{IEEEtran}
\bibliography{reference}

\end{document}